\documentclass[10pt,twocolumn,letterpaper]{article}

\usepackage[final]{cvpr}       

\usepackage{graphicx}
\usepackage{capt-of}
\usepackage{xcolor}
\usepackage{amsmath,amssymb,amsfonts,dsfont,pifont,bm,bbm,mathrsfs,mathtools,nicefrac}
\usepackage{algorithm,algorithmic,listings}
\usepackage{booktabs,multirow,adjustbox,diagbox,threeparttable,bbding}
\definecolor{citeblue}{RGB}{48,111,186}
\usepackage[pagebackref,breaklinks,colorlinks,citecolor=citeblue,bookmarks=false]{hyperref}
\usepackage[capitalize]{cleveref}  
\crefname{section}{Sec.}{Secs.}
\Crefname{section}{Section}{Sections}
\crefname{table}{Tab.}{Tabs.}
\Crefname{table}{Table}{Tables}
\crefname{figure}{Fig.}{Figs.}
\Crefname{figure}{Figure}{Figures}
\crefname{equation}{Eq.}{Eqs.}
\Crefname{equation}{Equation}{Equations}
\newcommand{\tocite}[1]{\textcolor{red}{[TO CITE]}}

\def\cvprPaperID{2739}
\def\confName{CVPR}
\def\confYear{2023}

\begin{document}
	
	\title{VideoFusion: Decomposed Diffusion Models for High-Quality Video Generation}
	
	\author{
		\footnotemark[1] Zhengxiong Luo\textsuperscript{1,2,4,5}\hspace{0.5em}
		Dayou Chen\textsuperscript{2}\hspace{0.5em}
		Yingya Zhang\textsuperscript{2}\hspace{0.5em}\\
		\footnotemark[2] Yan Huang\textsuperscript{4,5}\hspace{0.5em}
		Liang Wang\textsuperscript{4,5}\hspace{0.5em}
		Yujun Shen\textsuperscript{3}\hspace{0.5em}
		Deli Zhao\textsuperscript{2}\hspace{0.5em}
		Jingren Zhou\textsuperscript{2}\hspace{0.5em}
		Tieniu Tan\textsuperscript{4,5,6}\\
		\textsuperscript{1}University of Chinese Academy of Sciences (UCAS) \hspace{1em} \textsuperscript{2}Alibaba Group \\
		\textsuperscript{3}Ant Group \hspace{1em}
		\textsuperscript{4}Center for Research on Intelligent Perception and Computing (CRIPAC)\\
		\textsuperscript{5}Institute of Automation, Chinese Academy of Sciences (CASIA) \hspace{1em}
		\textsuperscript{6}Nanjing University\\
		{\tt\small zhengxiong.luo@cripac.ia.ac.cn \quad \{dayou.cdy, yingya.zyy, jingren.zhou\}@alibaba-inc.com}\\ {\tt\small \{shenyujun0302, zhaodeli\}@gmail.com\quad \{yhuang, wangliang, tnt\}@nlpr.ia.ac.cn}
	}
	
	\twocolumn[{%
		\renewcommand\twocolumn[1][]{#1}%
		\maketitle
		\begin{center}
			\centering
			\includegraphics[width=\linewidth]{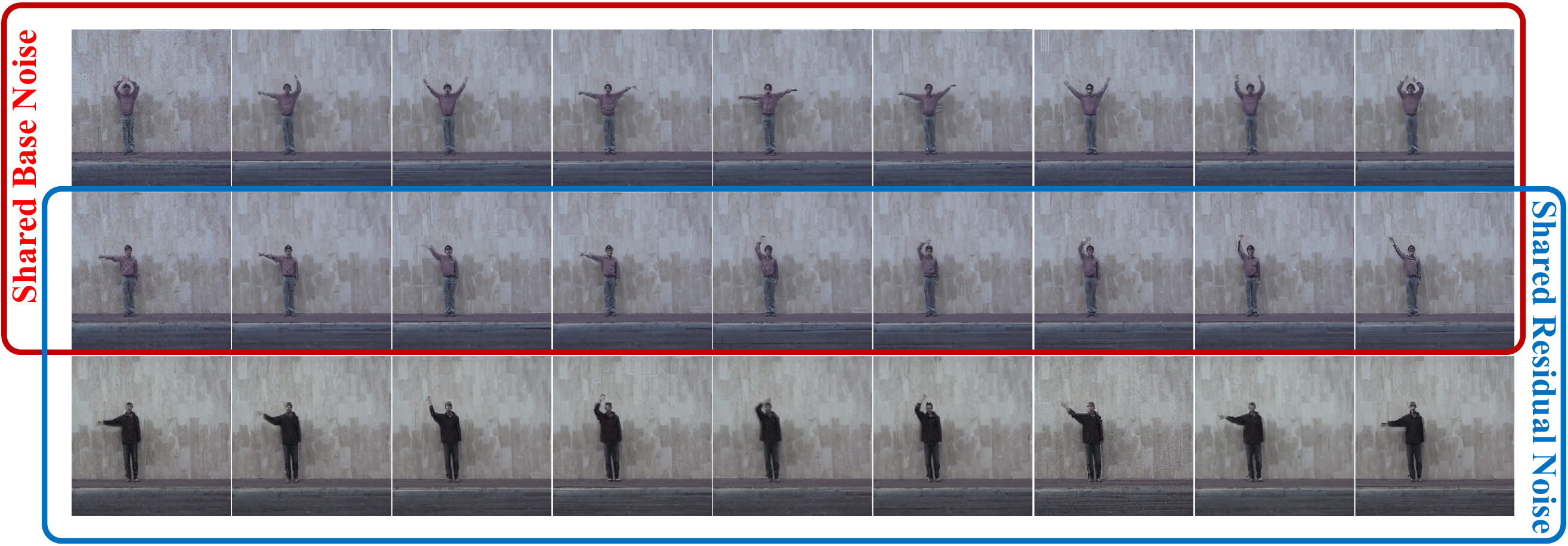}
			\vspace{-18pt}
			\captionsetup{type=figure}
			\caption{Unconditional generation results on the Weizmann Action dataset~\cite{human_actions}. Videos of the top-two rows share the same base noise but have different residual noises. Videos of the bottom two rows share the same residual noise but have different base noises.}\label{fig:human_actions}
			\vspace{8pt}
		\end{center}
	}]
	
	\footnotetext[1]{Work done at Alibaba DAMO Academy.}
	\footnotetext[2]{Corresponding author.}
	\begin{abstract}
		A diffusion probabilistic model (DPM), which constructs a forward diffusion process by gradually adding noise to data points and learns the reverse denoising process to generate new samples, has been shown to handle complex data distribution. Despite its recent success in image synthesis, applying DPMs to video generation is still challenging due to high-dimensional data spaces. Previous methods usually adopt a standard diffusion process, where frames in the same video clip are destroyed with independent noises, ignoring the content redundancy and temporal correlation. This work presents a decomposed diffusion process via resolving the per-frame noise into a base noise that is shared among all frames and a residual noise that varies along the time axis. The denoising pipeline employs two jointly-learned networks to match the noise decomposition accordingly. Experiments on various datasets confirm that our approach, termed as VideoFusion, surpasses both GAN-based and diffusion-based alternatives in high-quality video generation. We further show that our decomposed formulation can benefit from pre-trained image diffusion models and well-support text-conditioned video creation.
	\end{abstract}
	\vspace{-10pt}
	
	\section{Introduction}
	
	Diffusion probabilistic models (DPMs) are a class of  deep generative models, which consist of : \textit{i})  a diffusion process that gradually adds noise to data points, and \textit{ii}) a denoising process that generates new samples via iterative denoising~\cite{ddpm,fast-dpm}. Recently, DPMs have made awesome achievements in generating high-quality and diverse images~\cite{ddim, ddpm-var,glide,imagen,dalle-2,liu2022compositional}. 
	
	Inspired by the success of DPMs on image generation, many researchers are trying to apply a similar idea to video prediction/interpolation~\cite{fdm,mcvd,rvd}. While study about DPMs for video generation is still at an early stage~\cite{vdm} and faces challenges since video data are of higher dimensions and involve complex spatial-temporal correlations.
	
	Previous DPM-based video-generation methods usually adopt a standard diffusion process, where frames in the same video are added with independent noises and the temporal correlations are also gradually destroyed in noised latent variables. Consequently, the video-generation DPM is required to reconstruct coherent frames from independent noise samples in the denoising process. However, it is quite challenging for the denoising network to simultaneously model spatial and temporal correlations.
	
\begin{figure}[t]
		\centering
		\includegraphics[width=\linewidth]{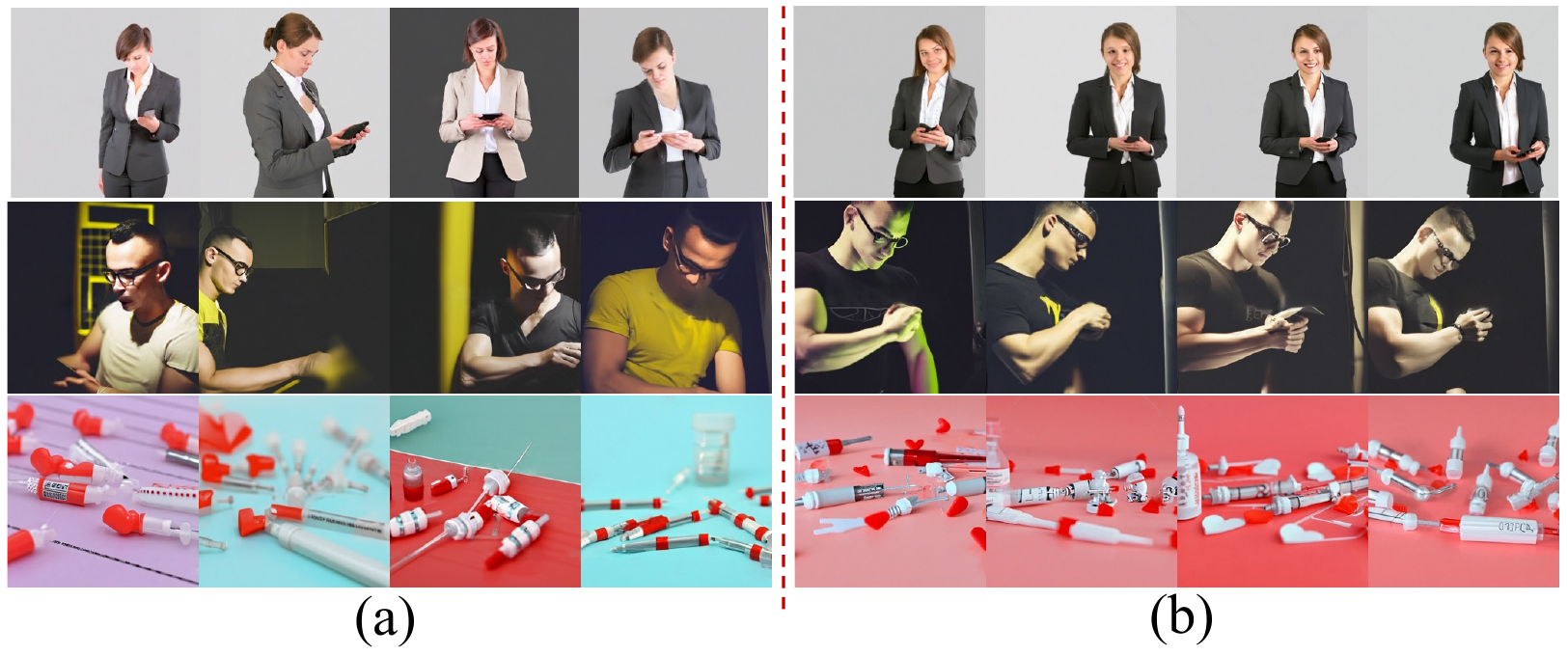}
		\vspace{-0.08\linewidth}
		\caption{Comparison between images generated from (a)~independent noises; (b)~noises with a shared base noise. Images of the same row are generated by the decoder of DALL-E 2~\cite{dalle-2} with the same condition.}\label{fig:share_vs_idd}
		\vspace{-12pt}
	\end{figure}

	Inspired by the idea that consecutive frames share most of the content, we are motivated to think: would it be easier to generate video frames from noises that also have some parts in common? To this end, we modify the standard diffusion process and propose a decomposed diffusion probabilistic model, termed as VideoFusion, for video generation. During the diffusion process, we resolve the per-frame noise into two parts, namely \textit{base noise} and \textit{residual noise}, where the base noise is shared by consecutive frames. In this way, the noised latent variables of different frames will always share a common part, which makes the denoising network easier to reconstruct a coherent video. For intuitive illustration, we use the decoder of  DALL-E 2~\cite{dalle-2} to generate images conditioned on the same latent embedding. As shown in~\cref{fig:share_vs_idd}a, if the images are generated from independent noises, their content varies a lot even if they share the same condition. But if the noised latent variables share the same base noise, even an image generator can synthesize roughly correlated sequences (shown in~\cref{fig:share_vs_idd}b). Therefore, the burden of the denoising network of video-generation DPM can be largely alleviated.
	
	Furthermore, this decomposed formulation brings additional benefits.
	Firstly, as the base noise is shared by all frames, we can predict it by feeding one frame to a large pretrained image-generation DPM with only one forward pass. In this way, the image priors of the pretrained model could be efficiently shared by all frames and thereby facilitate the learning of video data.
	Secondly, the base noise is shared by all video frames and is likely to be related to the video content. This property makes it possible for us to better control the content or motions of generated videos. Experiments in \cref{exp:decompose} show that, with adequate training, VideoFusion tends to relate the base noise with video content and the residual noise to motions (\cref{fig:human_actions}). 
	Extensive experiments show that VideoFusion can achieve state-of-the-art results on different datasets and also well support text-conditioned video creation.
	
	\section{Related Works}
	\subsection{Diffusion Probabilistic Models}
	DPM is first introduced in~\cite{ddpm-first}, which consists of a diffusion (encoding) process and a denoising (decoding) process. In the diffusion process, it gradually adds random noises to the data $\mathbf{x}$ via a $T$-step Markov chain~\cite{fast-dpm}. The noised latent variable at step $t$ can be expressed as: 
	\begin{equation}
		\mathbf{z}_t = \sqrt{\hat{\alpha}_t}x + \sqrt{1 - \hat{\alpha}_t} \mathbf{\epsilon_t}
	\end{equation}
	with
	\begin{equation}
		\hat{\alpha}_t=\prod \limits_{k=1}^t \alpha_k \quad \mathbf{\epsilon_t}\sim \mathcal{N}(\mathbf{0}, \mathbf{1}),
	\end{equation}
	where $\alpha_t\in(0, 1)$ is the corresponding diffusion coefficient. For a $T$ that is large enough, \eg $T=1000$, we have $\sqrt{\hat{\alpha}_T}\approx0$ and $\sqrt{1 - \hat{\alpha}_T}\approx1$. And $\mathbf{z}_T$ approximates a random gaussian noise. Then the generation of $\mathbf{x}$ can be modeled as iterative denoising.
	
	In~\cite{ddpm}, Ho~\etal connect DPM with denoising score matching~\cite{score-match} and propose a $\epsilon$-prediction form for the denoising process:
	\begin{equation}
		\mathcal{L}_t = \|\mathbf{\epsilon}_t - \mathbf{z}_{\mathbf{\theta}}(\mathbf{z}_t, t)\|^2,
	\end{equation}
	where $\mathbf{z}_{\mathbf{\theta}}$ is a denoising neural network parameterized by $\mathbf{\theta}$, and $\mathcal{L}_t$ is the loss function. Based on this formulation, DPM has been applied to various generative tasks, such as image-generation~\cite{classifier-free,dalle-2},  super-resolution~\cite{srdiff,sr_dpm}, image translation~\cite{unit-ddpm}, \etc, and become an important class of deep generative models. Compared with generative adversarial networks (GANs)~\cite{gan}, DPMs are easier to be trained and able to generate more diverse samples~\cite{dpm-beat-gan,latent-dpm}.
	
	\subsection{Video Generation}
	Video generation is one of the most challenging tasks in the generative research field. It not only needs to generate high-quality frames but also the generated frames need to be temporally correlated. Previous video-generation methods are mainly GAN-based. In VGAN~\cite{vgan} and TGAN~\cite{tgan}, the generator is directly used to learn the joint distribution of video frames. In~\cite{mocogan}, Tulyakov~\etal propose to decompose a video clip into content and motion and model them respectively via a content vector and a motion vector. A similar decomposed formulation is also adopted in~\cite{stylegan-v} and~\cite{nvidia-long}, in which the content noise is shared by consecutive frames to learn the video content and a motion noise is used to model the object trajectory. Other methods firstly train a vector quantized auto encoder~\cite{vqvae,vqgan,vqdpm,cogview,cogview2} for video data, and then use an auto-regress transformer~\cite{transformer} to learn the video distribution in the quantized latent space~\cite{tats,videogpt,cogvideo}.
	
	Recently, inspired by the great achievements of DPM in image generation, many researchers also try to apply DPM to video generation. In~\cite{vdm}, Ho~\etal propose a video diffusion model, which extends the 2D denoising network in image diffusion models to 3D by stacking frames together as the additional dimension. In~\cite{mcvd}, DPM is used for video prediction and interpolation with the known frames as the condition for denoising. However, these methods usually treat video frames as independent samples in the diffusion process, which may make it difficult for DPM to reconstruct coherent videos in the denoising process.
	
	\section{Decomposed Diffusion Probabilistic Model}
	\subsection{Standard Diffusion Process for Video Data}
	Suppose $\mathbf{x}=\{x^i \mid  i = 1, 2, \dots, N\}$ is a video clip with $N$ frames, and $\mathbf{z}_t = \{z^i_t \mid i = 1, 2, \dots, N\}$ is the noised latent variable of $\mathbf{x}$ at step $t$. Then the transition from $x^i$ to $z^i_t$ can be expressed as:
	\begin{equation}
		z^i_t = \sqrt{\hat{\alpha}_t}x^i + \sqrt{1 - \hat{\alpha}_t}\epsilon^i_t, \label{ori_diffusion}
	\end{equation}
	where $\epsilon^i_t \sim \mathcal{N}(0, 1)$.
	
	In previous methods, the added noise $\epsilon_t$ of each frame is independent of each other. And frames in the video clip $\mathbf{x}$ are encoded to $\mathbf{z_T}\approx\{\epsilon^i_T \mid i = 1, 2, \dots, N\}$, which are independent noise samples. This diffusion process ignores the relationship between video frames. Consequently, in the denoising process, the denoising network is expected to reconstruct a coherent video from these independent noise samples. Although this task could be realized by a denoising network that is powerful enough, the burden of the denoising network may be alleviated if the noise samples are already correlated. Then it comes to a question: can we utilize the similarity between consecutive frames to make the denoising process easier?
	
	\subsection{Decomposing the Diffusion Process}
	To utilize the similarity between video frames, we split the frame $x^i$ into two parts: a base frame $x^0$ and a residual $\Delta x^i$:
	\begin{equation}
		x^i = \sqrt{\lambda^i}x^0 + \sqrt{1 - \lambda^i}\Delta x^i, \quad i = 1, 2, ..., N
	\end{equation}
	where $x^0$ represents the common parts of the video frames, and  $\lambda^i \in [0, 1]$ represents the proportion of $x_0$ in $x^i$. Specially, $\lambda^i=0$ indicates that $x^i$ has nothing in common with $x^0$,  and $\lambda^i=1$ indicates $x^i = x^0$.  In this way, the similarity between video frames can be grasped via $x^0$ and $\lambda^i$. And the noised latent variable at step $t$ is:
	\begin{equation}
		z^i_t =\sqrt{\hat{\alpha}_t}(\sqrt{\lambda^i}x^0 + \sqrt{1 - \lambda^i}\Delta x^i) + \sqrt{1 - \hat{\alpha}_t}\epsilon^i_t .\label{split_x}
	\end{equation}
	
	Accordingly, we also split the added noise $\epsilon^i_t$ into two parts: a base noise $b^i_t$ and a residual noise $r^i_t$:
	\begin{equation}
		\epsilon^i_t = \sqrt{\lambda^i} b^i_t + \sqrt{1 - \lambda^i} r^i_t \quad b^i_t, r^i_t \sim \mathcal{N}(0, 1).\label{split_noise}
	\end{equation}
	We substitute Eq~\ref{split_noise} into~\cref{split_x} and get 
	\begin{equation}
		\begin{aligned}
			z^i_t 
			=\sqrt{\lambda^i}&(\underbrace{\sqrt{\hat{\alpha}_t}x^0 + \sqrt{1 - \hat{\alpha}_t}b^i_t)}_{\text{diffusion of } x^0} +\\
			\sqrt{1 - \lambda^i}&(\underbrace{\sqrt{\hat{\alpha}_t}\Delta x^i +  \sqrt{1 - \hat{\alpha}_t}r^i_t)}_{\text{diffusion of } \Delta x^i}. \label{split_diffusion}
		\end{aligned}
	\end{equation}
	
	\begin{figure}[t]
		\centering
		\includegraphics[width=\linewidth]{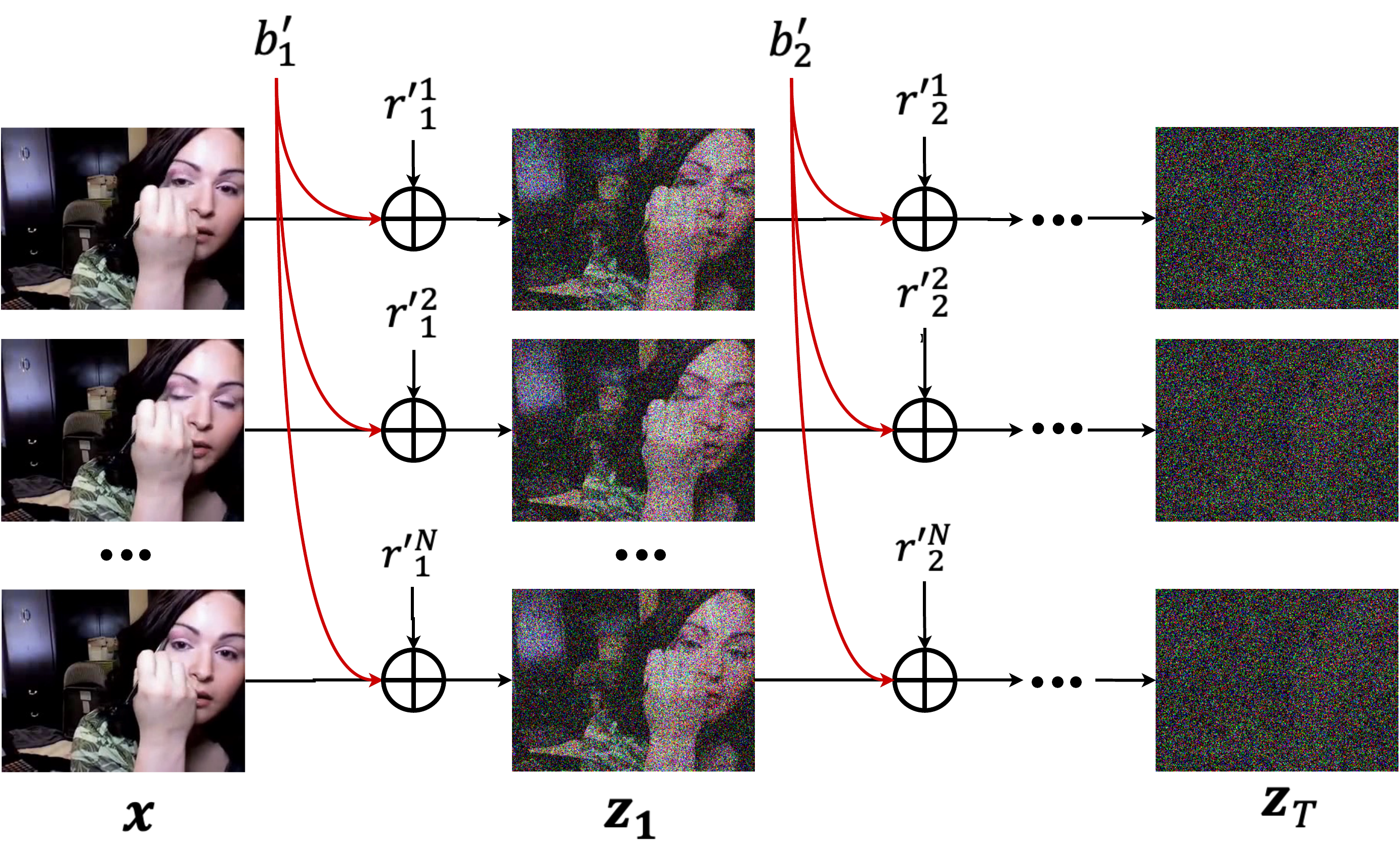}
		\caption{The decomposed diffusion process of a video clip. The base noise $b_t$ is shared across different frames. For simplicity, we omit the coefficient of each component in the figure.}\label{vis_diffusion}
	\end{figure}
	
	As one can see in~\cref{split_diffusion}, the diffusion process can be decomposed into two parts: the diffusion of $x^0$ and the diffusion of $\Delta x^i $. In previous methods, although $x^0$ is shared by consecutive frames, it is independently noised to different values in each frame, which may increase the difficulty of denoising. Towards this problem, we propose to share $b^i_t$ for $i=1, 2, ..., N$ such that $b^i_t = b_t$. In this way, $x^0$ in different frames will be noised to the same value. And frames in the video clip $\mathbf{x}$ will be encoded to $\mathbf{z}_T \approx \{\sqrt{\lambda^i}b_T + \sqrt{1 - \lambda^i}r^i_T\mid i = 1, 2, \dots, N\}$, which is sequence of noise samples correlated via $b_T$. From these samples, it may be easier for the denoising network to reconstruct a coherent video.
	
	With shared $b_t$, the latent noised variable $z^i_t$ can be expressed as:
	\begin{equation}
		z^i_t =\sqrt{\hat{\alpha}_t}x^i + \sqrt{1 - \hat{\alpha}_t}(\sqrt{\lambda^i}b_t + \sqrt{1-\lambda^i}r^i_t). \label{share_noise}
	\end{equation}
	As shown in~\cref{vis_diffusion}, this decomposed form also holds between adjacent diffusion steps:
	\begin{equation}
		z^i_t = \sqrt{\alpha_t}z^i_{t-1} + \sqrt{1 - \alpha_t}(\sqrt{\lambda^i}b_t^{\prime} + \sqrt{1-\lambda^i}r^{\prime i}_t),
	\end{equation}
	where $b_t^{\prime}$ and $r^{\prime i}_t$ are respectively the base noise and residual noise at step $t$. And $b_t^{\prime}$ is also shared between frames in the same video clip.
	
	\begin{figure*}[t]
		\centering
		\includegraphics[width=\linewidth]{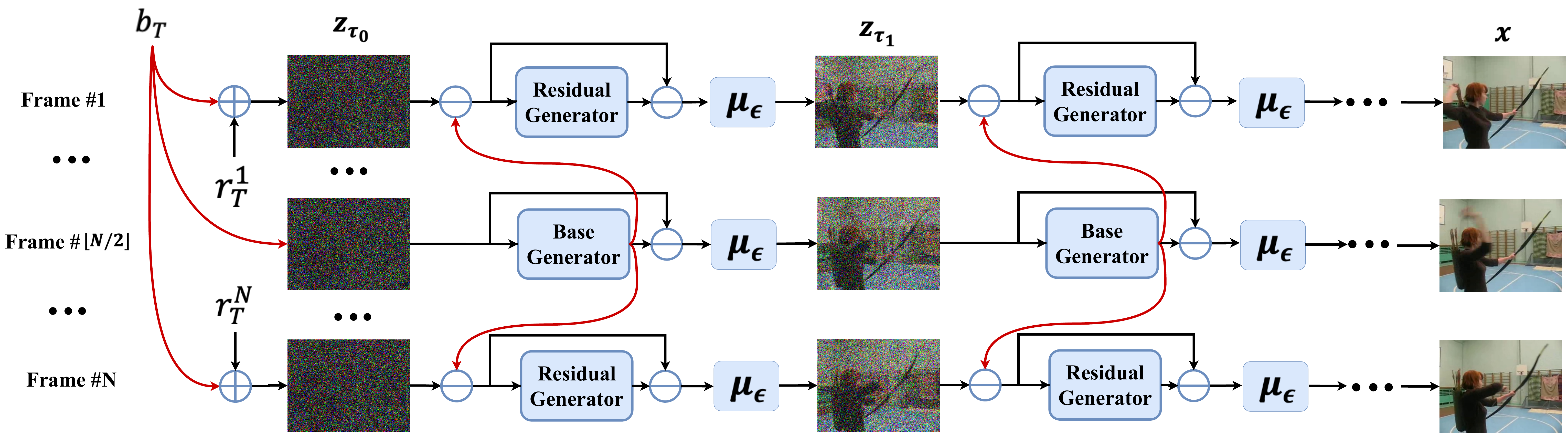}
		\caption{Visualization of DDIM~\cite{ddim} sampling process of VideoFusion. In each sampling step, we first remove the base noise with the base generator and then estimate the remaining residual noise via the residual generator. $\tau_i$ denotes the DDIM sampling steps. $\mu_{\epsilon}$ denotes mean-value predicted function of DDIM in $\epsilon$-prediction formulation. We omit the coefficients and conditions in the figure for simplicity.}\label{fig:vis_denoise}
	\end{figure*}
	
	\subsection{Using a Pretrained Image DPM}\label{sec:use_pretrain}
	Generally, for a video clip $\mathbf{x}$, there is an infinite number of choices for $x^0$ and $\lambda^i$ that satisfy~\cref{split_x}. But we hope $x^0$ contains most information of the video, \eg the background or main subjects of the video, such that $x^i$ only needs to model the small difference between $x^i$ and $x^0$. Empirically, we set $x^0=x^{\lfloor N/2 \rfloor}$ and  $\lambda^{\lfloor N/2 \rfloor} = 1$, where $\lfloor \cdot \rfloor$ denotes the floor rounding function. In this case , we have $\Delta x^{\lfloor N/2 \rfloor} = 0$ and~\cref{share_noise} can be simplified as:
	\begin{equation}
		\begin{aligned}
			& z^i_t = \\
			& \left\{
			\begin{aligned}
				&\sqrt{\hat{\alpha}_t}x^i + \sqrt{1 - \hat{\alpha}_t} b_t \quad i = \lfloor N/2 \rfloor\\
				&\sqrt{\hat{\alpha}_t}x^i + \sqrt{1 - \hat{\alpha}_t}(\sqrt{\lambda^i}b_t + \sqrt{1-\lambda^i}r^i_t) \quad i \neq \lfloor N/2 \rfloor,
			\end{aligned}
			\right.
		\end{aligned}\label{middle_diffusion}
	\end{equation}
	
	We notice that~\cref{middle_diffusion} provides us a chance to estimate the base noise $b_t$ for all frames with only one forward pass by feeding $x^{\lfloor N/2 \rfloor}$ into a $\epsilon$-prediction denoising function $\mathbf{z}^{b}_{\mathbf{\phi}}$ (parameterized by $\mathbf{\phi}$). We call $\mathbf{z}^{b}_{\mathbf{\phi}}$ as the \textit{base generator}, which is a denoising network of an image diffusion model. It enables us to use a pretrained image generator, \eg DALL-E 2~\cite{dalle-2} and Imagen~\cite{imagen}, as the base generator. In this way, we can leverage the image priors of the pretrained image DPM, thereby facilitating the learning of video data.
	
	As shown in~\cref{fig:vis_denoise}, in each denoising step, we first estimate the based noised as $z^{b}_{\mathbf{\phi}}(z^{\lfloor N/2 \rfloor}_t, t)$, and then remove it from all frames:
	\begin{equation}
		z^{\prime i}_{t} = z^i_t -  \sqrt{\lambda^i}\sqrt{1 - \hat{\alpha}_t}z^{b}_{\mathbf{\phi}}(z^{\lfloor N/2 \rfloor}_t, t)
		\quad
		i \neq {\lfloor N/2 \rfloor}.\label{eq:z_prime}
	\end{equation}
	We then feed $z^{\prime i}_{t}$ into a \textit{residual generator}, denoted as $\mathbf{z}^{r}_{\mathbf{\psi}}$ (parameterized by $\mathbf{\psi}$), to estimate the residual noise $r^i_t$ as $\mathbf{z}^{r}_{\mathbf{\psi}}(z^{\prime i}_{t}, t, i)$. We need to note that the residual generator is conditioned on the frame number $i$ to distinguish different frames. As $b_t$ has already been removed, $z^{\prime i}_{t}$ is expected to be less noisy than $z^i_t$. Then it may be easier for $\mathbf{z}^{r}_{\mathbf{\psi}}$ the estimate the remaining residual noise. According to~\cref{split_noise} and~\cref{middle_diffusion}, the noise $\epsilon^i_t$ can be predicted as:
	\begin{equation}
		\begin{aligned}
			& \left\{
			\begin{aligned}
				&  \mathbf{z}^{b}_{\mathbf{\phi}}(z^{\lfloor N/2 \rfloor}_t, t)  \quad i ={\lfloor N/2 \rfloor} \\
				& \sqrt{\lambda^i}\mathbf{z}^{b}_{\mathbf{\phi}}(z^{\lfloor N/2 \rfloor}_t, t) + \sqrt{1-\lambda^i} \mathbf{z}^{r}_{\mathbf{\psi}}(z^{\prime i}_{t}, t, i)  \quad i \neq {\lfloor N/2 \rfloor},
			\end{aligned}
			\right.
		\end{aligned}\label{eq:predited_noise}
	\end{equation}
	where $z^{\prime i}_{t}$ can be calculated by~\cref{eq:z_prime}. Then, we can follow the denoising process of DDIM~\cite{ddim} (shown in~\cref{fig:vis_denoise}) or DDPM~\cite{ddpm} (shown in Appendix) to infer the next latent diffusion variable and loop until we get the sample~$x^i$.
	
	As indicated in~\cref{eq:predited_noise}, the base generator $\mathbf{z}^b_{\mathbf{\phi}}$ is responsible for reconstructing the base frame $x^{\lfloor N/2 \rfloor}$, while $\mathbf{z}^r_{\mathbf{\psi}}$ is expected to reconstruct the residual $\Delta x^i$. Often, $x^0$ contains rich details and is difficult to be learned. In our method, a pretrained image-generation model is used to reconstruct $x^0$, which largely alleviates this problem. Moreover, in each denoising step, $\mathbf{z}^b_{\mathbf{\phi}}$ takes in only one frame, which allows us to use a large pretrained model (up to $2$-billion parameters) while consuming an affordable graph memory. Compared with $x^0$, the residual $\Delta x^i$ may be much easier to be learned~\cite{rvd,masked_flow,agustsson2020scale,yang2020hierarchical}. Therefore, we can use a relatively smaller network (with $0.5$-billion parameters) for the residual generator. In this way, we concentrate more parameters on the more difficult task, \ie the learning of $x^0$, and thereby improve the efficiency of the whole method.
	
	\subsection{Joint Training of Base and Residual Generators}\label{sec:joint_training}
	In ideal cases, the pretrained base generator can be kept fixed during the training of VideoFusion. However, we experimentally find that fixing the pretrained model will lead to unpleasant results. We attribute this to the domain gap between the image data and video data. Thus it is helpful to simultaneously finetune the base generator $\mathbf{z}^b_{\mathbf{\theta}}$ on the video data with a small learning rate. We define the final loss function as:
	\begin{equation}
		\small
		\begin{aligned}
			&\mathcal{L}_t = \\
			&\left\{
			\begin{aligned}
				& \| \epsilon^i_t -  \mathbf{z}^{b}_{\mathbf{\phi}}(z^{\lfloor N/2 \rfloor}_t, t)\|^2  \quad i ={\lfloor N/2 \rfloor} \\
				& \| \epsilon^i_t - \sqrt{\lambda^i}[\mathbf{z}^{b}_{\mathbf{\theta}}(z^{\lfloor N/2 \rfloor}_t, t)]_{sg} - \sqrt{1-\lambda^i} \mathbf{z}^{r}_{\mathbf{\mathbf{\psi}}}(z^{\prime i}_{t}, t, i)\|^2  \quad i \neq {\lfloor N/2 \rfloor},
			\end{aligned}
			\right.
		\end{aligned}
	\end{equation}
	where $[\cdot]_{sg}$ is the \textit{stop-gradient} operation, which means that the gradients will not be propagated back to $\mathbf{z}^b_{\mathbf{\theta}}$ when $i \neq {\lfloor N/2 \rfloor}$. We hope that the pretrained model is finetuned only by the loss on the base frame. This is because at the beginning of the training, the estimated results of  $\mathbf{z}^{r}_{\mathbf{\psi}}(z^{\prime i}_{t}, t)$ is noisy which may destroy the pretrained model.
	
	\subsection{Discussions}\label{sec:decompose}
	In some GAN-based methods, the videos are generated from two concatenated noises, namely \textit{content code} and \textit{motion code}, where the content code is shared across frames~\cite{mocogan,stylegan-v}. These methods show the ability to control the video content (motions) by sampling different content (motion) codes. It is difficult to directly apply such an idea to DPM-based methods, because the noised latent variables in DPM should have the same shape as the generated video. In the proposed VideoFusion, we decompose the added noise by representing it as the weighted sum of base noise and residual noise, in which way, the latent video space can also be decomposed. According to the DDIM sampling algorithm in~\cref{vis_diffusion}, the shared base frame $x^{{\lfloor N/2 \rfloor}}$ is only dependent on the base noise $b_T$. It enables us to control the video content via $b_T$, \eg generating videos with the same content but different motions by keeping $b_T$ fixed, which helps us generate longer coherent sequences in~\cref{sec:exp_long_seq}. But it may be difficult for VideoFusion to automatically learn to relate the residual noise to video motions, as it is difficult for the residual generator to distinguish the base or residual noises from their weighted sum. Whereas in~\cref{exp:decompose}, we experimentally find if we provide VideoFusion with explicit training guidance that videos with the same motions in a mini-batch also share the same residual noise, VideoFusion could also learn to relate the residual noise to video motions.
	
	\section{Experiments}
	\subsection{Experimental Setup}
	\noindent
	\textbf{Datasets.}
	For quantitative evaluation, we train and test our method on three datasets, \ie UCF101~\cite{ucf}, Sky Time-lapse~\cite{sky}, and TaiChi-HD~\cite{taichi}. On UCF101, we show both unconditional and class-conditioned generation results. while on  Sky Time-lapse and TaiChi-HD, only unconditional generation results are provided. For quantitative evaluation, we also train a text-conditioned video-generation model on WebVid-10M~\cite{webvid}, which consists of 10.7M short videos with paired textual descriptions.
	
	\vspace{0.02\linewidth}\noindent
	\textbf{Metrics.}
	Following previous works~\cite{cogvideo,tats}, we mainly use Fr\'echet Video Distance (FVD)~\cite{fvd}, Kernel Video Distance (KVD)~\cite{fvd}, and Inception Score (IS)~\cite{is} as the evaluation metrics. We use the evaluation scripts provided in~\cite{tats}. All metrics are evaluated on videos with $16$ frames and $128\times128$ resolution. On UCF101, we report the results of IS and FVD, and on Sky Time-lapse and TaiChi-HD we report the results of FVD and KVD.
	
	\vspace{0.02\linewidth}\noindent
	\textbf{Training.}
	We use a pretrained decoder of DALL-E 2~\cite{dalle-2} (trained on Laion-5B~\cite{laion5b}) as our base generator, while the residual generator is a randomly initialized  2D U-shaped denoising network~\cite{imagen,ddpm}. In the training phase, both the base generator and the residual generator are conditioned on the image embedding extracted by the visual encoder of CLIP~\cite{clip} from the central image of the video sample. A \textit{prior} is also trained to generate latent embedding. For conditional video generation, the prior is conditioned on video captions or classes. And for unconditional generation, the condition of the prior is empty text.
	Our models are initially trained on $16$-frame video clips with $64\times64$ resolution and then super-resolved to higher resolutions with DPM-based SR models~\cite{sr_dpm}. Without special statement, we set $\lambda^i=0.5, \forall i \neq 8$ and $\lambda^8=0.5$.
	
	\begin{table}[t]
		\centering
		\caption{Quantitative comparisons on UCF101.$\downarrow$ denotes the lower the better. $\uparrow$ denotes the higher the better. The best results are denoted in bold.}\label{table:ucf_compare}
		\vspace{-5pt}
		\setlength{\tabcolsep}{0.4cm}
		\resizebox{\linewidth}{!}{
			\begin{tabular}{llll}
				\toprule
				Method           & Resolution & IS $\uparrow$      & FVD$\downarrow$    \\
				\midrule
				\multicolumn{4}{l}{\textit{Unconditional}}\\
				\midrule
				TGAN~\cite{tgan}
				&$16\times64\times64$& $11.85$ & $-$    \\
				MoCoGAN-HD~\cite{mocogan}
				&$16\times128\times128$& $32.36$ & $838$  \\
				DIGAN~\cite{digan}            
				&$16\times128\times128$& $32.70$ & $577$  \\
				StyleGAN-V~\cite{stylegan-v} 
				&$16\times256\times256$& $23.94$ & $-$ \\
				VideoGPT~\cite{videogpt}
				&$16\times128\times128$& $24.69$ & $-$    \\
				TATS~\cite{tats}
				&$16\times128\times128$& $57.63$ & $420$ \\
				VDM~\cite{vdm}
				&$16\times64\times64$& $57.00$ & $295$ \\
				VideoFusion
				&$16\times64\times64$& $71.67$ & $\mathbf{139}$ \\
				VideoFusion
				&$16\times128\times128$& $\mathbf{72.22}$ & $220$ \\
				\midrule
				\multicolumn{4}{l}{\textit{Class-conditioned}}\\
				\midrule
				VGAN~\cite{vgan}  
				&$16\times64\times64$& $8.31$  & $-$    \\
				TGAN~\cite{tgan} 
				&$16\times64\times64$& $15.83$  & $-$    \\
				TGANv2~\cite{is}
				&$16\times128\times128$& $28.87$ & $1209$ \\
				MoCoGAN~\cite{mocogan}
				&$16\times64\times64$& $12.42$ & $-$    \\
				DVD-GAN~\cite{dvd-gan}
				&$16\times128\times128$& $32.97$ & $-$    \\
				CogVideo~\cite{cogvideo}
				&$16\times160\times160$& $50.46$ & $626$ \\
				TATS~\cite{tats}
				&$16\times128\times128$& $79.28$ & $332$ \\
				VideoFusion
				&$16\times128\times128$& $\mathbf{80.03}$ & $\mathbf{173}$ \\
				\bottomrule
		\end{tabular}}
		\vspace{-12pt}
	\end{table}
	
	\subsection{Quantitative Results}
	We compare our methods with several competitive methods, including VideoGPT~\cite{videogpt}, CogVideo~\cite{cogvideo}, StyleGAN-V~\cite{stylegan-v},  DIGAN~\cite{digan}, TATS~\cite{tats}, VDM~\cite{vdm}, \etc.  Most of these methods are GAN-based except that VDM is DPM-based. The quantitative results on UCF101 of these methods are shown in~\cref{table:ucf_compare}. On unconditional generations, VDM outperforms the GAN-based methods in the table, especially in terms of  FVD. It implies the potential of DPM-based video-generation methods. While compared with VDM, the proposed VideoFusion further outperforms it by a large margin on the same resolution. The superiority of VideoFusion may be attributed to the more appropriate diffusion framework and a strong pretrained image DPM. If we increase the resolution to $16\times128\times128$, the IS of VideoFusion can be further improved. We notice that the FVD score will get worse when VideoFusion generates videos with a higher resolution. This is possible because videos with higher resolutions contain richer details and are more difficult to be learned. Nevertheless, VideoFusion achieves the best quantitative results on UCF101. We also provide the FVD and KVD results (with resolution as $16\times128\times128$) on Sky Time-lapse and TaiChi-HD in~\cref{table:sky_compare} and~\cref{table:taichi_compare} respectively. As one can see, VideoFusion still achieves much better results than previous methods.
	
	\begin{table}[t]
		\centering
		\caption{Quantitative comparisons on Sky Time-lapse~\cite{sky}. $\downarrow$ denotes the lower the better. The best results are denoted in bold.}\label{table:sky_compare}
		\vspace{-5pt}
		\setlength{\tabcolsep}{0.7cm}
		\resizebox{\linewidth}{!}{
			\begin{tabular}{lcc}
				\toprule
				Method & FVD ($\downarrow$) & KVD ($\downarrow$) \\
				\midrule
				MoCoGAN-HD~\cite{mocogan}
				& $183.6$ & $13.9$ \\
				DIGAN~\cite{digan}
				& $114.6$ & $6.8$ \\
				TATS~\cite{tats}
				& $132.6$ & $5.7$ \\
				\midrule
				VideoFusion &$\mathbf{47.0}$&$\mathbf{5.3}$ \\
				\bottomrule
		\end{tabular}}
	\end{table}
	\begin{table}[t]
		\centering
		\caption{Quantitative comparisons on TaiChi-HD~\cite{taichi}. $\downarrow$ denotes the lower the better. The best results are denoted in bold.}\label{table:taichi_compare}
		\vspace{-5pt}
		\setlength{\tabcolsep}{0.7cm}
		\resizebox{\linewidth}{!}{
			\begin{tabular}{lcc}
				\toprule
				Method & FVD ($\downarrow$) & KVD ($\downarrow$) \\
				\midrule
				MoCoGAN-HD~\cite{mocogan}
				& $144.7$ & $25.4$ \\
				DIGAN~\cite{digan}
				& $128.1$ & $20.6$ \\
				TATS~\cite{tats}
				& $94.6$ & $9.8$ \\
				\midrule
				VideoFusion
				&$\mathbf{56.4}$& $\mathbf{6.9}$\\
				\bottomrule
		\end{tabular}}
	\end{table}
	
	\begin{figure*}[t]
		\centering
		\includegraphics[width=\linewidth]{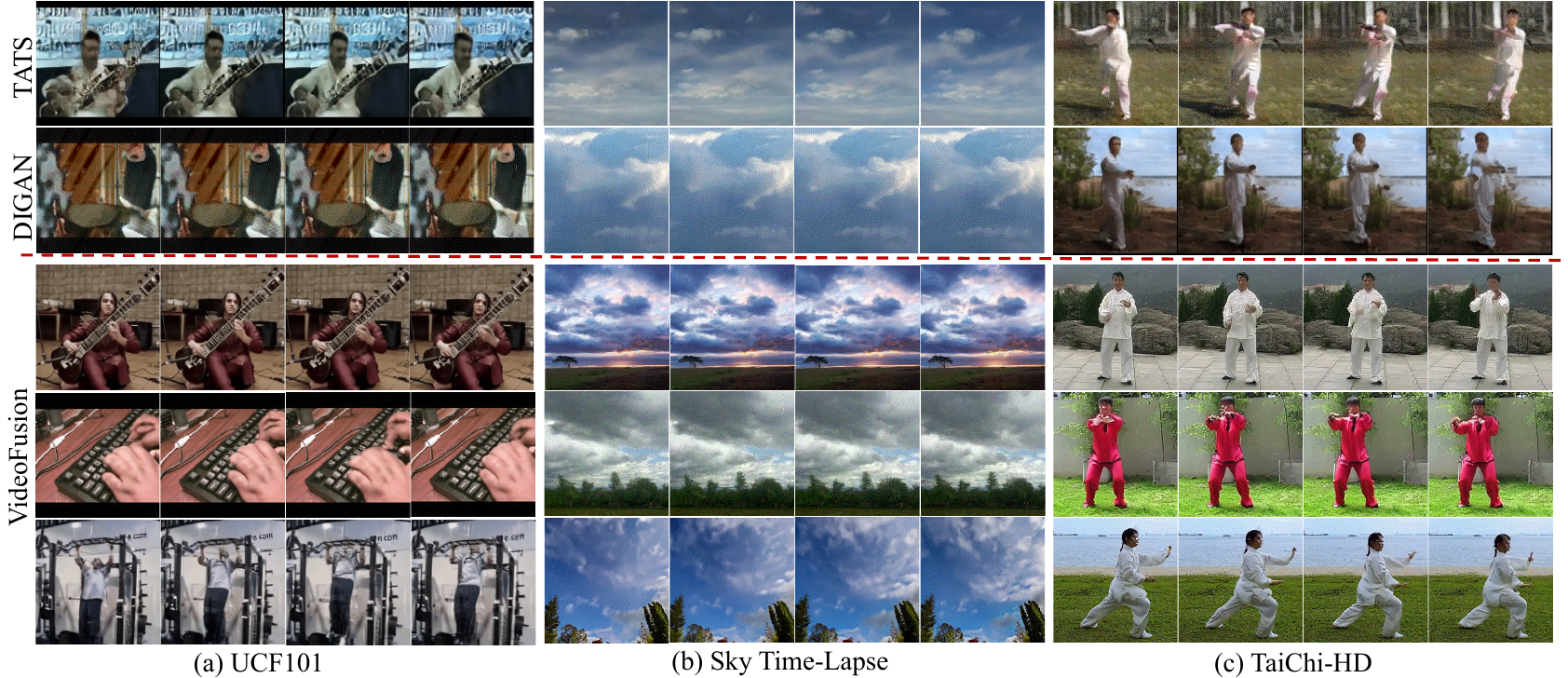}
		\vspace{-20pt}
		\caption{Visual comparisons on (a) UCF101~\cite{ucf}, (b) Sky Time-Lapse~\cite{sky},  and (c) TaiChi-HD~\cite{taichi}.}\label{fig:vis_compare}
		\vspace{-10pt}
	\end{figure*}
	
	\subsection{Qualitative Results}
\begin{figure*}[t]
		\centering
		\includegraphics[width=\linewidth]{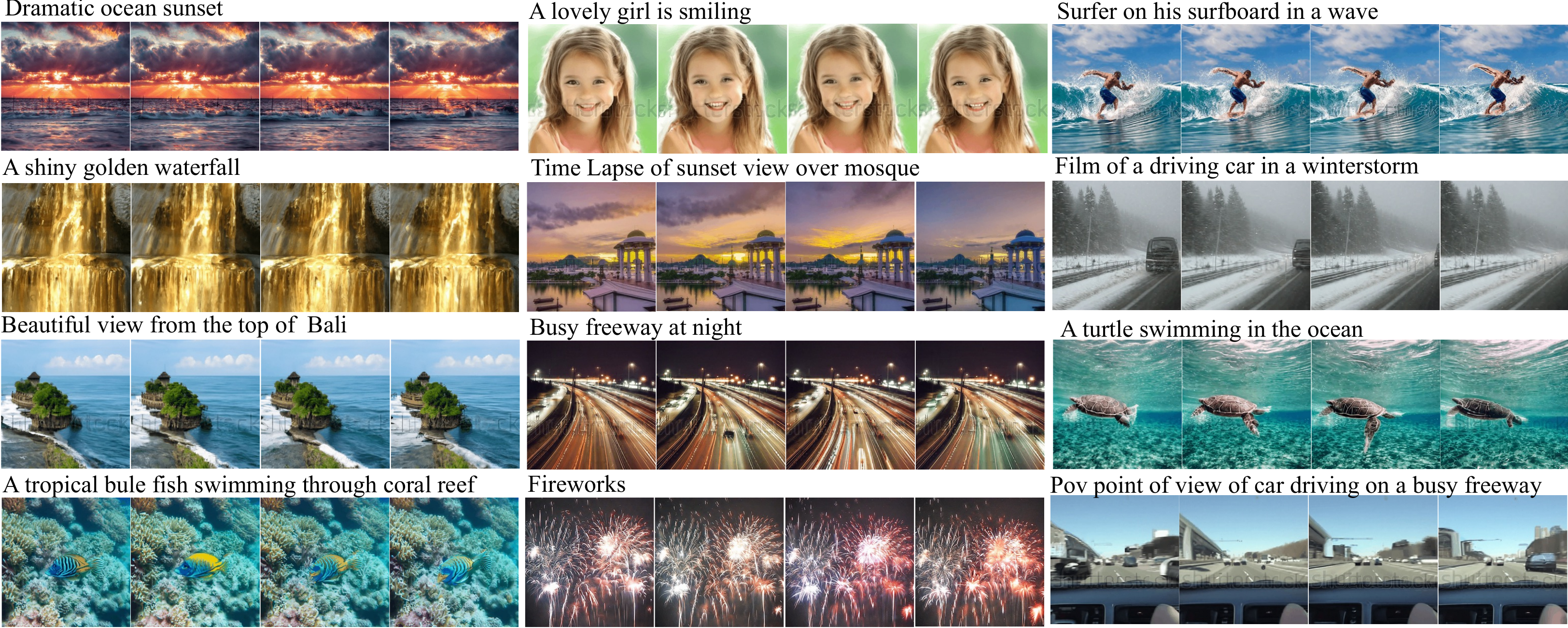}
		\vspace{-12pt}
		\caption{Text-to-video generation results of  VideoFusion trained on WebVid-10M~\cite{webvid}.}\label{fig:webvid}
		\vspace{-15pt}
	\end{figure*}

	We also provide visual comparisons with the most recent state-of-the-art methods, \ie TATS and DIGAN. As shown in~\cref{fig:vis_compare}, each generated video has $16$ frames with a resolution of $128\times128$ and we show the $4^{th}$, $8^{th}$, $12^{th}$ and $16^{th}$ frame in the figure. As one can see, our VideoFusion can generate more realistic videos with richer details.  To further demonstrate the quality of videos generated by VideoFusion, we train a text-to-video model on the large-scale video dataset, \ie WebVid-10M. Some samples are shown in~\cref{fig:webvid}.
	
	\subsection{Efficiency Comparison}
	As we have discussed in~\cref{sec:use_pretrain}, in each denoising step VideoFusion estimates the base noise for all frames with only one forward pass. It allows us to use a large base generator while keeping the computational cost affordable. As a comparison, previous video-generation DPM, \ie VDM~\cite{vdm}, extends a 2D DPM to 3D by stacking images at an additional dimension, which processes each frame in parallel and may introduce redundant computations. To make a quantitative comparison, we re-implement VDM based on the base generator of VideoFusion. We evaluate the inference memory and speed of VideoFusion and VDM in~\cref{tab:eff}. As one can see, despite that  VideoFusion consists of an additional residual generator and prior, its consumed memory is reduced by  $\mathbf{21.8\%}$ and latency is reduced by $\mathbf{57.5\%}$ when compared with VDM.  This is because the powerful pretrained base generator allows us to use a smaller residual generator, and the shared base noise requires only one forward pass of the base generator.
	
	\begin{figure}[t]
		\centering
		\includegraphics[width=\linewidth]{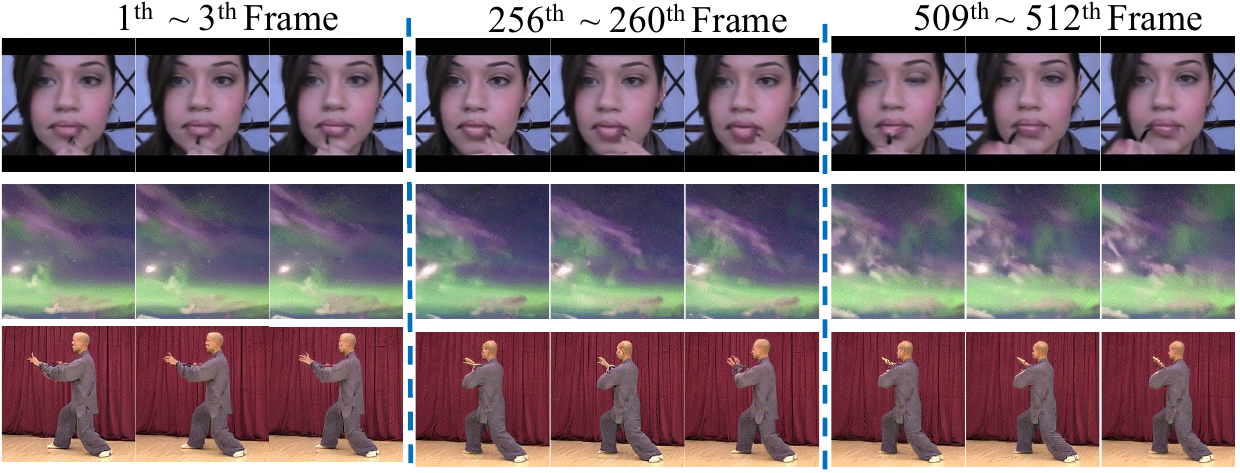}
		\vspace{-18pt}
		\caption{Generated videos with $512$ frames on UCF101~\cite{ucf} (top-row), Sky Time-Lapse~\cite{sky} (second-row), and TaiChi-HD\cite{taichi} (third-row).}\label{fig:long_sequence}
		\vspace{-12pt}
	\end{figure}

	\subsection{Ablation Study}
	\noindent
	\textbf{Study on $\lambda^i$.}
	To explore the influence of $\lambda^i, \forall i\neq \lfloor N/2 \rfloor$, we perform controlled experiments on UCF101. As one can see in~\cref{tab:lambda}, if  $\lambda^i$ is too small, \eg $\lambda^i=0.1$, or $\lambda^i$ is too large, \eg $\lambda^i = 0.75$, the performance of VideoFusion will get worse. A small $\lambda^i$ indicates that less base noise is shared across frames, which makes it difficult for VideoFusion to exploit the temporal correlations. While a large $\lambda^i$ suggests that the video frames share most of their information, which restricts the dynamics in the generated videos.  Consequently, an appropriate $\lambda^i$ is important for VideoFusion to achieve better performance.
	
	\begin{table}[t]
		\centering
		\caption{We re-implement VDM~\cite{vdm} (denoted as VDM$^{*}$) based on the base generator of VideoFusion. The efficiency comparisons are shown below.}\label{tab:eff}
		\vspace{-5pt}
		\setlength{\tabcolsep}{0.7cm}
		\resizebox{\linewidth}{!}{
			\begin{tabular}{lll}
				\toprule
				Method & Memory (GB) & Latency (s) \\
				\midrule
				VDM$^{*}$ & $63.82$ & $0.40$\\
				VideoFusion & $49.85(\downarrow 21.8\%)$ & $0.17(\downarrow 57.5\%)$ \\
				\bottomrule
		\end{tabular}}
	\end{table}
	\begin{table}[t]
		\centering
		\caption{Study on $\lambda^i$. Unconditional generation results on UCF101~\cite{ucf}.}\label{tab:lambda}
		\vspace{-8pt}
		\setlength{\tabcolsep}{0.5cm}
		\resizebox{\linewidth}{!}{
			\begin{tabular}{lcccc}
				\toprule
				$\lambda^i$ & $0.10$ & $0.25$ & $0.50$ & $0.75$ \\
				\midrule
				IS $\uparrow$&$67.23$& $69.16$ & $71.67$&$69.56$ \\
				FVD $\downarrow$ & $149$& $122$  &$139$ & $181$\\
				\bottomrule
		\end{tabular}}
		\vspace{-10pt}
	\end{table}
	
	\vspace{0.02\linewidth}\noindent
	\textbf{Study on pretraining.}
	To further explore the influence of the pretrained model, we also train our VideoFusion from the scratch. The quantitative comparisons of unconditional generation results on UCF101 are shown in~\cref{tab:pretrain}. As one can see, a well-pretrained base generator does help VideoFusion achieve better performance, because the image priors of the pretrained model can ease the difficulty of learning the image content and help VideoFusion focus on exploiting the temporal correlations. We need to note that VideoFusion still surpasses VDM without the pretrained initialization. It also suggests the superiority of  VideoFusion against only come from the pretrained model, but also the decomposed formulation.
	
	\vspace{0.02\linewidth}\noindent
	\textbf{Study on joint training.}
	As we have discussed in~\cref{sec:joint_training}, we finetune the pretrained generator jointly with the residual generator. We experimentally compare different training methods in~\cref{tab:joint_training}. As one can see, if the pretrained base generator is fixed, the performance of VideoFusion is poor. This is because of the domain gap between the pretraining dataset (Laion-5B) and UCF101. The fixed pretrained base generator may provide misleading information on UCF101 and impede the training of the residual generator. If the base generator is jointly finetuned on UCF101, the performance will be largely improved. Whereas if we remove the stop-gradient technique mentioned in~\cref{sec:joint_training}, the performance gets worse. This is because the randomly initialized residual generator would destroy the pretrained model at the beginning of training. 
	
	\subsection{Generating Long Sequences}\label{sec:exp_long_seq}
	Limited by computational resources, a video-generation model usually generates only a few frames in one forward pass. Previous methods mainly adopt an auto-regressive framework for generating longer videos~\cite{tats,videogpt}. However, it is difficult to guarantee the content coherence of extended frames. In~\cite{song2020score}, Yang~\etal propose a \textit{replacement method} for DPMs to extend the generated sequences in an auto-regressive way. Whereas it still often fails to keep the content of extended video frames~\cite{vdm}. In our proposed method, the incoherence problem may be largely alleviated, since we can keep the base noise fixed when generating extended frames. To verify this idea, we use the replacement method to extend a $16$-frame video to $512$ frames. As shown in~\cref{fig:long_sequence}, both the quality and coherence can be well-kept in the extended frames.
	
	\begin{table}[t]
		\centering
		\caption{Study on pretraining. Unconditional generation results on UCF101~\cite{ucf}.}\label{tab:pretrain}
		\setlength{\tabcolsep}{0.8cm}
		\vspace{-5pt}
		\resizebox{\linewidth}{!}{
			\begin{tabular}{lll}
				\toprule
				Method & IS$\uparrow$& FVD $\downarrow$ \\
				\midrule
				VDM~\cite{vdm} & $57.00$ & $295$\\
				VideoFusion w/o pretrain & $65.29$& $183$ \\
				VideoFusion w/ pretrain & $71.67$& $139$  \\
				\bottomrule
		\end{tabular}}
	\end{table}
	\begin{table}[t]
		\centering
		\caption{Study on joint training. Unconditional generation results on UCF101~\cite{ucf}.}\label{tab:joint_training}
		\vspace{-5pt}
		\setlength{\tabcolsep}{0.8cm}
		\resizebox{\linewidth}{!}{
			\begin{tabular}{lcc}
				\toprule
				Training method & IS$\uparrow$& FVD $\downarrow$ \\
				\midrule
				Fixed  &$65.06$ & $187$\\
				w/o stop gradient & $67.86$& $168$ \\
				w stop gradient & $71.67$& $139$  \\
				\bottomrule
		\end{tabular}}
		\vspace{-10pt}
	\end{table}
	
	\subsection{Decomposing the Motion and Content}\label{exp:decompose}
	To further explore the ability of VideoFusion on decomposing the video motion and content, we perform experiments on the Weizmann Action dataset~\cite{human_actions}. It contains $81$ videos of $9$ people performing 9 actions, including jumping-jack and waving-hands \etc. As we have discussed in~\cref{sec:decompose}, it may be difficult for VideoFusion to automatically learn to correlate the residual noise with video motions. Thus, we provide VideoFusion with explicit training guidance. To this end, in each training mini-batch of VideoFusion, we share the base noise across videos with the same human identity and share residual noise across videos with the same actions.  Since the difference between frames of the same video in Weizmann Action dataset is relatively small, we set $\lambda^i = 0.9, \forall i \neq \lfloor N/2\rfloor$ in this experiment. The generated results are shown in~\cref{fig:human_actions}. As one can see, by keeping the base noise fixed and sampling different residual noises, VideoFusion succeeds to keep the human identity in videos of different actions. Also, by keeping the residual noise fixed and sampling different base noises, VideoFusion generates videos of the same action but with different human identities.
	
	\section{Limitations and Future Work}
	Sharing base noise among consecutive frames helps the video-generation DPM to better exploit the temporal correlation, however, it may also limit motions in the generated videos. Although we can adjust $\lambda^i$ to control the similarity between consecutive frames, it is difficult to find a suitable $\lambda^i$ for all videos, since in some videos the differences between frames is small, while in other videos the differences may be large. In the future, we will try to adaptively generate $\lambda^i$ for each video and even each frame. 
	
	And in the current version of VideoFusion, the residual generator is conditioned on the latent embedding produced by the pretrained prior of DALL-E 2. This is because the embedding condition can help the residual generator converge faster. This practice works well for unconditional generations or generations from relatively short texts,  in which cases the latent embedding may be enough for encoding all conditioning information. Whereas in video generation from long texts, it may be difficult for the prior to encode the long temporal information of the caption into the latent embedding. A better way is to condition the residual generator directly on the long text. However, the modality gap between the text data and video data will largely increase the burden of the residual generator and make it difficult for the residual generator to converge. In the future, we may try to alleviate this problem and explore video generation from long texts.
	
	\section{Conclusion}
	In this paper, we present a decomposed DPM for video generation (VideoFusion). It decomposes the standard diffusion process as adding a base noise and a residual noise, where the base noise is shared by consecutive frames. In this way, frames in the same video clip will be encoded to a correlated noise sequence, from which it may be easier for the denoising network to reconstruct a coherent video. Moreover, we use a pretrained image-generation DPM to estimate the base noise for all frames with only one forward pass, which leverages the priors of the pretrained model efficiently. Both quantitative and qualitative results show that VideoFusion can produce results competitive to the-state-of-art methods.  
	
	\section*{Acknowledgment}
	This work was jointly supported by National Natural Science Foundation of China (62236010, 62276261,61721004, and U1803261), Key Research Program of Frontier Sciences CAS Grant No. ZDBS-LYJSC032, Beijing Nova Program (Z201100006820079), and CAS-AIR.
	
	{\small
		\bibliographystyle{ieee_fullname}
		\bibliography{ref}
	}
	
	\newpage
	\appendix
	\section{DDPM Sampling of VideoFusion}
	The DDPM sampling algorithm of VideoFusion is shown in~\cref{alg:ddpm}. We need note that during each sampling process, the added noise is also resolved into a base noise and a residual noise, where the base noise is shared across frames and residual noise varies along time axis.
	
	\section{Details about VideoFusion}
	
	The base generator and residual generator are both U-shape networks. For experiments on UCF101~\cite{ucf}, Sky Time-lapse~\cite{sky}, and TaiChi-HD~\cite{taichi}, we use relative smaller models. The details are shown in~\cref{tab:small}. For experiments on the large-scale datasets, \ie WebVid-10M~\cite{webvid}, we use relatively large models, whose details are shown in~\cref{tab:large}. As one can see, we use a large pretrained base generator ($2.00$ billion parameters) on WebVid-10M. Since the knowledge of  the pretrained model can be efficiently shared by all frames via its predicted base noise, we can use a  smaller residual generator ($0.59$ billion parameters) to save the computations.
	
	\begin{algorithm}[t]
		\caption{DDPM~\cite{ddpm} sampling of VideoFusion} 
		\label{alg:ddpm}
		\begin{algorithmic}
			\STATE Sampling $b\sim \mathcal{N}(0, 1)$, $z^{ \lfloor N/2 \rfloor}\gets b$
			\FOR{$i=1$ to $N$ \AND $i\neq \lfloor N/2 \rfloor$}
			\STATE Sampling $r^i\sim \mathcal{N}(0, 1)$
			\STATE $z^i =\sqrt{\lambda^i} b + \sqrt{1 - \lambda^i} r^i$
			\ENDFOR
			\FOR{$t=T$ to $1$}
			\STATE $b \gets \mathbf{z}^b_{\mathbf{\phi}}(z^{\lfloor N/2 \rfloor}, t)$; $\epsilon^{\lfloor N/2 \rfloor} \gets b$
			\FOR{$i=1$ to $N$ \AND $i\neq \lfloor N/2 \rfloor$}
			\STATE $z^{\prime i} \gets z^i -  \sqrt{\lambda^i}\sqrt{1 - \hat{\alpha}}b$
			\STATE $r^i \gets \mathbf{z}^r_{\mathbf{\psi}}(z^{\prime i}, t, i)$
			\STATE $\epsilon^i \gets \sqrt{\lambda^i} b + \sqrt{1 - \lambda^i} r^i$
			\ENDFOR
			\FOR{$i=1$ to $N$}
			\STATE $\mu^i \gets \frac{1}{\sqrt{\alpha_t}}z^i - \frac{1-\alpha_t}{\sqrt{1-\hat{\alpha}_t}\sqrt{\alpha_t}}\epsilon^i$
			\STATE $\sigma \gets \frac{1-\hat{\alpha}_{t-1}}{1-\hat{\alpha}_t}(1 - \alpha_t)$
			\ENDFOR
			\STATE Sampling $b\sim \mathcal{N}(0, 1)$
			\STATE $z^{\lfloor N/2 \rfloor}\gets \sigma b + \mu^{\lfloor N/2 \rfloor}$
			\FOR{$i=1$ to $N$ \AND $i\neq \lfloor N/2 \rfloor$}
			\STATE Sampling $r^i\sim \mathcal{N}(0, 1)$
			\STATE $z^i \gets \sigma(\sqrt{\lambda^i} b + \sqrt{1 - \lambda^i} r^i) + \mu^i $
			\ENDFOR
			\ENDFOR
			\RETURN $\{z^i\mid i = 1, 2, \dots, N\}$
		\end{algorithmic} 
	\end{algorithm}
	
	\begin{table}[t]
		\centering
		\caption{Details about the base generator and residual generator on UCF101~\cite{ucf}, Sky Time-Lapse~\cite{sky}, and TaiChi-HD~\cite{taichi}.}\label{tab:small}
		\setlength{\tabcolsep}{0.45cm}
		\resizebox{\linewidth}{!}{
			\begin{tabular}{lll}
				\toprule
				&Base generator&Residual  generator\\
				\midrule
				\multicolumn{3}{l}{\textit{Network}} \\
				\midrule
				base dims
				& $192$ & $128$ \\
				dim expansions
				&$1, 2, 3, 5$ &$1, 2, 3, 5$ \\
				Textual embedding dims
				& $768$ & $768$ \\
				Visual embedding dims
				& $768$ & $768 $ \\ 
				\# Scales
				& $4$ &$4$ \\
				\# Layers per scale
				& $2$ & $2$\\
				Attention head dims
				& $64$  & $64$ \\
				Attention scales
				&$\frac{1}{2}, \frac{1}{4}, \frac{1}{8}$&$\frac{1}{2}, \frac{1}{4}, \frac{1}{8}$ \\
				\# Params (B)
				& $0.29$ & $0.22$ \\
				\midrule
				\multicolumn{3}{l}{\textit{Diffusion}}\\
				\midrule
				Diffusion steps
				& $1000$ & $1000$ \\
				Noise schedule
				& cosine &cosine \\
				Sampling algorithm
				& DDIM & DDIM \\
				Sampling steps
				& $50$ & $50$ \\
				Variance type
				& fixed small & fixed small \\
				Guidance
				& classifier free & classifier free \\
				Guiding scale
				& $3.0$ & $3.0$ \\
				Objective
				& eps & eps \\
				Loss type
				& MSE & MSE\\
				\bottomrule
		\end{tabular}}
	\end{table}
	\begin{table}[t]
		\centering
		\caption{Details about the base generator and residual generator on WebVid-10M~\cite{webvid}.}\label{tab:large}
		\setlength{\tabcolsep}{0.45cm}
		\resizebox{\linewidth}{!}{
			\begin{tabular}{lll}
				\toprule
				&Base generator&Residual  generator\\
				\midrule
				\multicolumn{3}{l}{\textit{Network}} \\
				\midrule
				base dims
				& $512$ & $192$ \\
				dim expansions
				&$1, 2, 3, 4$ &$1, 2, 3, 4$ \\
				Textual embedding dims
				& $768$ & $768$ \\
				Visual embedding dims
				& $768$ & $768 $ \\ 
				\# Scales
				& $4$ &$4$ \\
				\# Layers per scale
				& $3$ & $3$\\
				Attention head dims
				& $64$  & $64$ \\
				Attention scales
				&$\frac{1}{2}, \frac{1}{4}, \frac{1}{8}$&$\frac{1}{2}, \frac{1}{4}, \frac{1}{8}$ \\
				\# Params (B)
				& $2.00$ & $0.59$ \\
				\midrule
				\multicolumn{3}{l}{\textit{Diffusion}}\\
				\midrule
				Diffusion steps
				& $1000$ & $1000$ \\
				Noise schedule
				& cosine &cosine \\
				Sampling algorithm
				& DDIM & DDIM \\
				Sampling steps
				& $50$ & $50$ \\
				Variance type
				& fixed small & fixed small \\
				Guidance
				& classifier free & classifier free \\
				Guiding scale
				& $3.0$ & $3.0$ \\
				Objective
				& eps & eps \\
				Loss type
				& MSE & MSE \\
				\bottomrule
		\end{tabular}}
	\end{table}
	
\end{document}